\documentclass{article}

\usepackage[final]{corl_2026} 

\usepackage{multicol}
\usepackage{multirow}
\usepackage{amssymb}   
\usepackage{amsfonts}  
\usepackage{amsmath}
\usepackage{bm}
\usepackage{booktabs}
\usepackage{pifont}
\usepackage{graphicx}
\usepackage{subcaption}
\usepackage{makecell}
\usepackage{bbm}
\usepackage{algorithm}
\usepackage{algpseudocode}
\usepackage{wrapfig}
\usepackage{xcolor}         

\definecolor{cmured}{HTML}{C41230}

\author{Yuxuan Kuang, Sungjae Park, Katerina Fragkiadaki$^{\dagger}$, Shubham Tulsiani$^{\dagger}$
\\ 
Carnegie Mellon University
\\
$^{\dagger}$ Equal Advising
\\
\href{https://dex4d.github.io}{\textcolor{cmured}{\texttt{https://dex4d.github.io}}}
}

\begin{document}

\title{{\method: Task-Agnostic Point Track Policy for Zero-Shot Dexterous Manipulation}}

\newcommand{\method}{\texttt{Dex4D}} 
\newcommand{\representation}{Paired Point Encoding} 
\newcommand{\yesmark}{\ding{51}} 
\newcommand{\nomark}{\ding{55}} 

\newcommand{\yuxuan}[1]{\textcolor{red}{\textbf{[Yuxuan]:} #1}}
\newcommand{\sj}[1]{\textcolor{blue}{\textbf{[sungjae]:} #1}}
\newcommand{\katef}[1]{\textcolor{purple}{\textbf{[Katerina]:} #1}}
\newcommand{\shubham}[1]{\textcolor{orange}{\textbf{[Shubham]:} #1}}

\maketitle



\begin{figure}[h]
\centering
\includegraphics[width=1.0\linewidth]{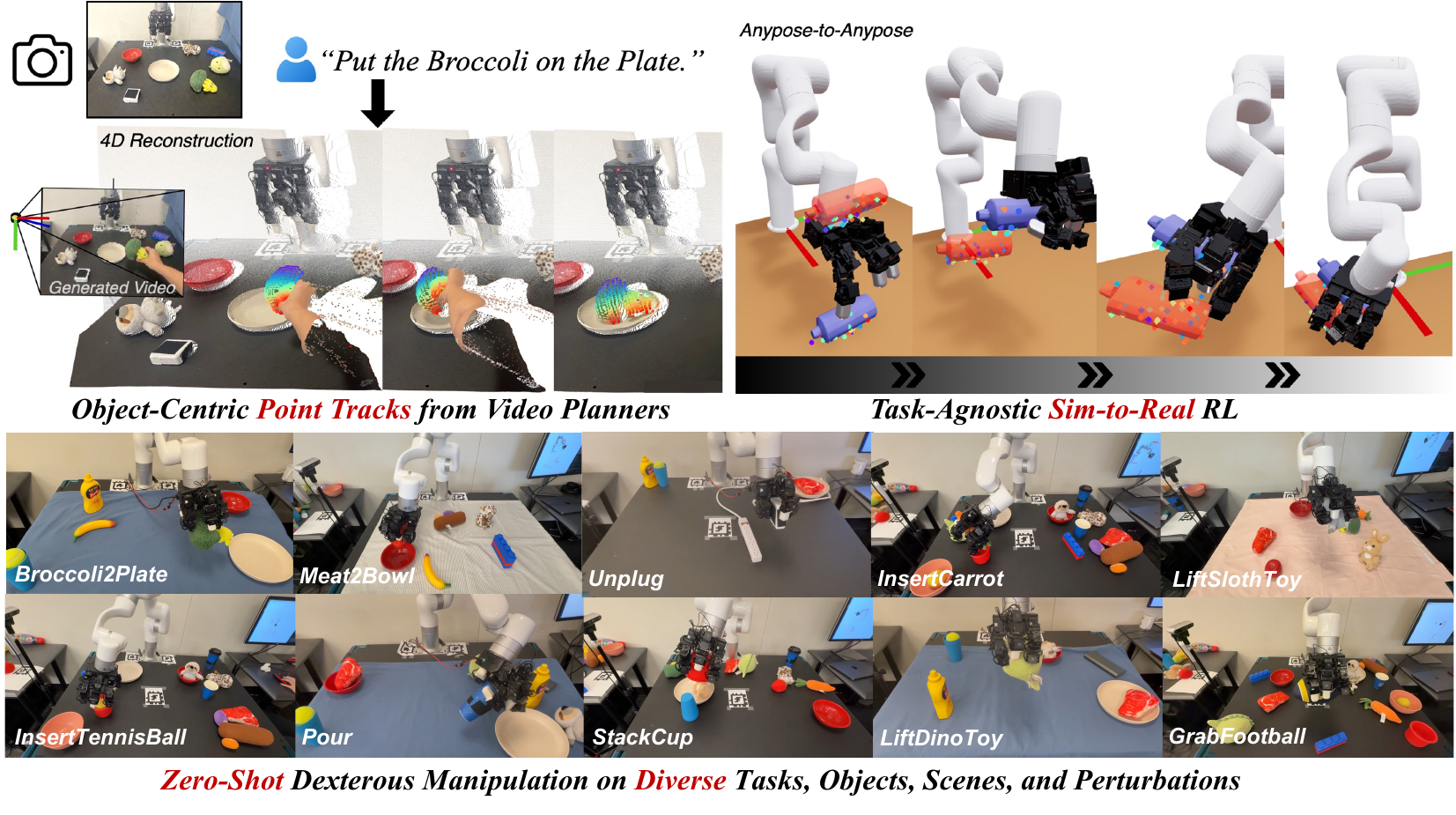}
  \caption{Overview of~\method. We leverage video generation and 4D reconstruction to generate \textbf{object-centric point tracks}. Conditioned on the point tracks, we use a \textbf{task-agnostic sim-to-real policy} trained via Anypose-to-Anypose for task execution. Our policy trained entirely in simulation can be seamlessly deployed in the real world \textbf{zero-shot} and generalizes to diverse configurations.}
  \label{fig:teaser}
\end{figure}
\begin{abstract}

Learning generalist policies capable of accomplishing a wide range of everyday tasks remains an open challenge in dexterous manipulation. In particular, collecting large-scale manipulation data via real-world teleoperation is expensive and difficult to scale. While learning in simulation provides a feasible alternative, designing multiple task-specific environments and rewards for training is similarly challenging. We propose \method{}, a framework that instead leverages simulation for learning \textbf{task-agnostic} dexterous skills that can be flexibly recomposed to perform diverse real-world manipulation tasks. Specifically, \method{} learns a domain-agnostic \textbf{3D point track conditioned} policy capable of transforming a rigid object from a current pose to a target pose. We train this `Anypose-to-Anypose' policy in simulation across thousands of objects with diverse pose configurations, covering a broad space of robot-object interactions that can be composed at test time.
At deployment, this policy can be zero-shot transferred to real-world tasks without finetuning, simply by prompting it with desired object-centric point tracks extracted from generated videos. During execution, \method{} uses online point tracking for closed-loop perception and control.
Extensive experiments in simulation and on real robots show that our method enables \textbf{zero-shot} deployment for diverse dexterous manipulation tasks and yields consistent improvements over prior baselines. Furthermore, we demonstrate strong generalization to novel objects, scene layouts, backgrounds, and trajectories, highlighting the robustness and scalability of the proposed framework. Project page: \texttt{\href{https://dex4d.github.io/}{\textcolor{cmured}{https://dex4d.github.io/}}}.

\end{abstract}

\keywords{Dexterous Manipulation, Reinforcement Learning}

\section{Introduction}
\label{sec:intro}


The lack of high-quality, diverse, and scalable data remains a fundamental bottleneck in learning dexterous robot manipulation. Collecting real-world manipulation trajectories is expensive, difficult to instrument, and limited in coverage and diversity. Furthermore, learning dexterous manipulation via teleoperation poses unique challenges due to the difficulty of precisely controlling high-dimensional robotic hands and fingers, which makes large-scale data collection slow and error-prone~\citep{yin2025dexteritygen}.
Learning dexterous manipulation behaviors via sim-to-real reinforcement learning (RL) provides a promising alternative \cite{xu2023unidexgrasp, chen2023bi}. Benefiting from highly parallel GPU-based simulation~\cite{makoviychuk2021isaac, mittal2025isaac} that largely improves interaction data bandwidth, RL-based policies can be learned in a few hours in simulation that are equivalent to years in the real world. However, training language-instructable `generalist' robot policies in simulation requires substantial engineering effort, including designing complex simulation environments, specifying task descriptions and instructions, performing tedious reward shaping, and tuning RL pipelines across an ever-growing set of tasks \cite{viral,robogen}.



We argue that instead of learning language-conditioned and task-specific policies, we can use highly parallel simulation to learn fundamental \textbf{task-agnostic} manipulation skills that can be flexibly composed using a high-level planner, such as video generation models~\cite{wiedemer2025veo3, wan2025wan} that have shown remarkable open-world generalization, to perform general downstream tasks. We operationalize this insight in our framework \method{}, which learns a \textbf{point track conditioned} policy for Anypose-to-Anypose -- \textit{manipulating any rigid object from any current pose to any target pose}. A key technical contribution lies in our goal representation -- instead of separately encoding current and target object points, we propose \textbf{\representation{}} that leverages the correspondences across them.  


We train our Anypose-to-Anypose policy across \textbf{thousands of objects} in simulation. The training process spans a broad space of object poses, trajectories, and hand-object interactions, enabling compositional generalization at test time. We show that our Anypose-to-Anypose policy, with geometry-aware and domain-robust point track representation as conditions, can be combined with scalable planners like video generation models to allow sim-to-real dexterous manipulation for generic tasks. Specifically, given a task description, \method{} queries a foundational video model to generate a successful video task plan. We then leverage 4D reconstruction to extract object-centric point tracks from the generated video as an interface for goal specifications and policy conditions for our task-agnostic Anypose-to-Anypose policy, while using efficient online point tracking for closed-loop perception and control. As a result, \method{} enables \textbf{zero-shot} transfer to real-world tasks \textbf{without any real robot finetuning}.

We evaluate \method{} extensively in simulation and on real robotic platforms. Comparing against state-of-the-art baselines, our method achieves a 87.5\% relative performance gain in real-world settings, showing substantial improvement and superiority over prior works. Furthermore, we demonstrate \textbf{strong generalization} to novel objects, scenes, tasks, and external perturbations, highlighting the scalability and robustness of the proposed representation and learning framework.

\section{Related Works}

\paragraph{Generalizable Dexterous Manipulation}

Endowing robots with human-level and generalizable dexterity is a long-standing goal for generalist robots that work under diverse real-world scenarios. It's also very challenging due to its high-DoF and high-dynamics nature. Prior optimization-based works often rely on contact-based optimization~\cite{liu2020deep, liu2021synthesizing, wang2022dexgraspnet, zhang2024dexgraspnet, yin2025lightning} to synthesize dexterous grasping poses, which are executed by motion planning. 
However, these works are mainly limited to grasping and prone to disturbances without closed-loop feedback. 
Another line of works uses mocap devices or teleoperation to collect dexterous manipulation data and train policies on them via imitation learning~\cite{wang2024dexcap, lin2025learning, wen2025gr}. 
However, these works suffer from in-domain data collection and fail to generalize to unseen tasks, objects, and scenes. 
Recently, reinforcement learning (RL) has shown promise on generalizable dexterous manipulation, including dexterous grasping~\cite{xu2023unidexgrasp, wan2023unidexgrasp++, lum2024dextrahg, singh2024dextrahrgb, singh2025end}, in-hand reorientation~\cite{qi2023hand, qi2023general, yin2025dexteritygen, wang2024lessons, liu2025dexndm}, and motion tracking~\cite{luo2024omnigrasp, liu2025dextrack, xu2025dexplore, patel2022learning}. 
Nonetheless, they often lack autonomy for high-level tasks that require task-specific planning. 
Recent work SimToolReal~\cite{kedia2026simtoolreal} also formulates a generalized hand controller as object-centric RL. The differences lie in the interface and its coverage: we use sparse point tracks for both observation and reward, so no object mesh or grasp bounding box is needed at test time, and we focus on a wider range of objects, spanning 3,200 objects rather than a small set of elongated tool primitives.
In contrast to all these works, our method leverages video generation and 4D reconstruction for high-level planning and trains an object-centric task-agnostic policy that works across tasks using sim-to-real RL with point tracks as an interface, achieving generalizable and autonomous dexterous manipulation.

\paragraph{Video-Based Robot Learning}

Recent years witnessed huge progress in video generation~\cite{brooks2024sora, wiedemer2025veo3, wan2025wan} and learning from human videos~\cite{bharadhwaj2024towards, bharadhwaj2024track2act, kuang2024ram, xu2024flow, bharadhwaj2024gen2act, park2025demodiffusion, wang2025crisp, wang2026counterfactual}. Video generation models not only can be used for entertainment or simulation, but also serve as world models or powerful high-level planners for robotics tasks since they are trained on enormous amounts of Internet videos and contain rich human priors~\cite{mei2026video}, therefore serving as a powerful system 2 (e.g., in Helix~\cite{figure2026introducing}). Recent works~\cite{patel2025robotic, zhi20253dflowaction, li2025novaflow, chen2025large, dharmarajan2025dream2flow} leverage video generation models or flow models as planners and use either pose estimation with motion planning or heuristic retargeting to map generated pixels to actions. However, these works suffer from large embodiment gaps and lack closed-loop feedback, which are crucial for highly dynamic tasks such as dexterous manipulation. They also require either object mesh~\cite{patel2025robotic} or clean point tracks~\cite{li2025novaflow} for pose estimation, which is hard to satisfy in the real world, especially with finger occlusions. In contrast, we train a closed-loop policy via sim-to-real, leveraging our proposed \representation~representation along with extensive point masking and domain randomization. Therefore, our method is robust to real-world noisy sensor input and can generalize to diverse unseen configurations.



\begin{figure*}[t]
\centering
\includegraphics[width=1.0\linewidth]{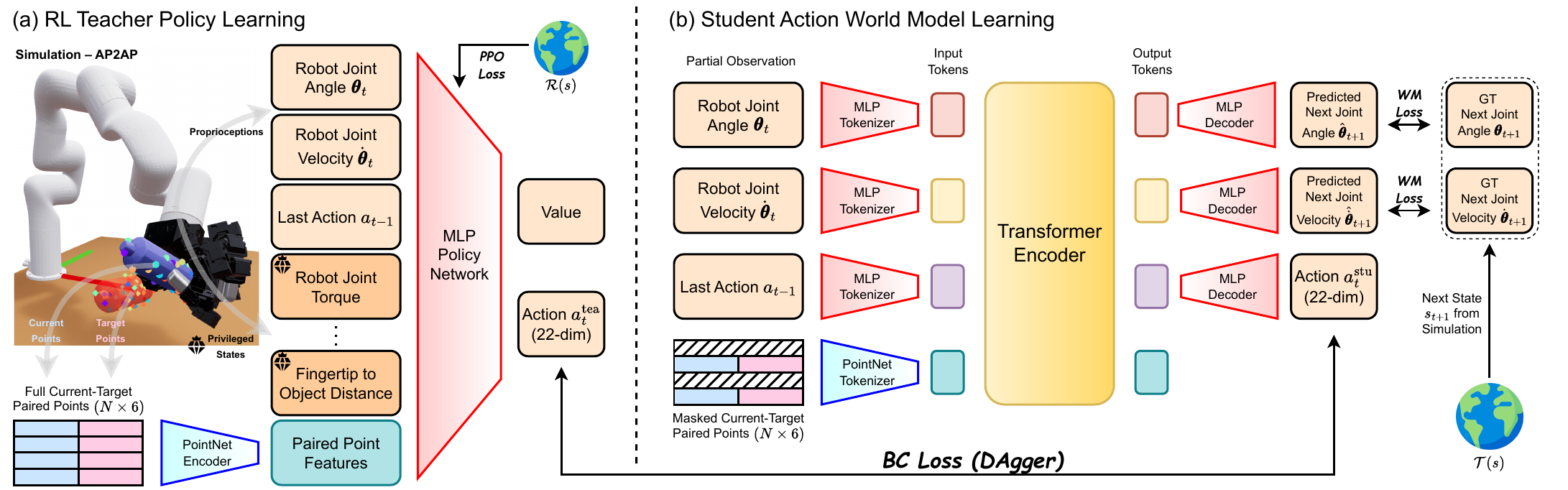}
  \caption{Overview of our~\method{} teacher and student network architectures. \textbf{(a)} We first learn a teacher policy via RL with privileged states and full points sampled on the whole object, leveraging our proposed~\textbf{\representation}~representation. \textbf{(b)} Given partial observation, \textit{i.e.,} robot proprioception, last action, and masked paired points, we distill the teacher to learn a \textbf{transformer-based student dynamics action model} that jointly predicts actions and future robot states.}
  \label{fig:method}
\end{figure*}
\section{Method}

In this section, we detail our Anypose-to-Anypose (AP2AP) setup (\S~\ref{sec:ap2ap-setup}), our proposed \representation~(\S~\ref{sec:representation} and Fig.~\ref{fig:representation}), teacher-student policy learning (\S~\ref{sec:teacher-student} and Fig.~\ref{fig:method}), and then outline how to deploy the sim-to-real AP2AP policy using point tracks from video generation planners in \S~\ref{sec:whole-deployment}.



\subsection{Anypose-to-Anypose}
\label{sec:ap2ap-setup}

Anypose-to-Anypose (AP2AP) is a task-agnostic sim-to-real learning formulation for dexterous manipulation. AP2AP abstracts manipulation as directly transforming an object from an arbitrary initial pose to an arbitrary target pose in 3D space, without assuming task-specific structure, predefined grasps, or motion primitives. Unlike prior approaches~\cite{patel2025robotic, li2025novaflow} that decompose manipulation into grasp generation, pose estimation, and planning, AP2AP treats object pose transformation itself as the fundamental learning objective, enabling a unified and reactive high-DoF control policy.


At the beginning of each episode, an object is placed on the table with a random initial position and orientation. The first goal requires the robot to grasp and lift the object to a specified pose. Once a goal is stably achieved, the next goal is randomly set as a nearby target pose, encouraging continuous pose-to-pose transitions and effective local exploration.
Our AP2AP policy is trained entirely in simulation using \textbf{3,200 objects} from UniDexGrasp~\cite{xu2023unidexgrasp} under diverse pose configurations and extensive domain randomization.
As a result, our method doesn't require task-specific tuning and generalizes zero-shot to unseen objects and downstream manipulation tasks in the real world.

\subsection{Goal Representation via~\representation}
\label{sec:representation}

\begin{wrapfigure}{r}{0.4\textwidth}
\centering
\includegraphics[width=1.0\linewidth]{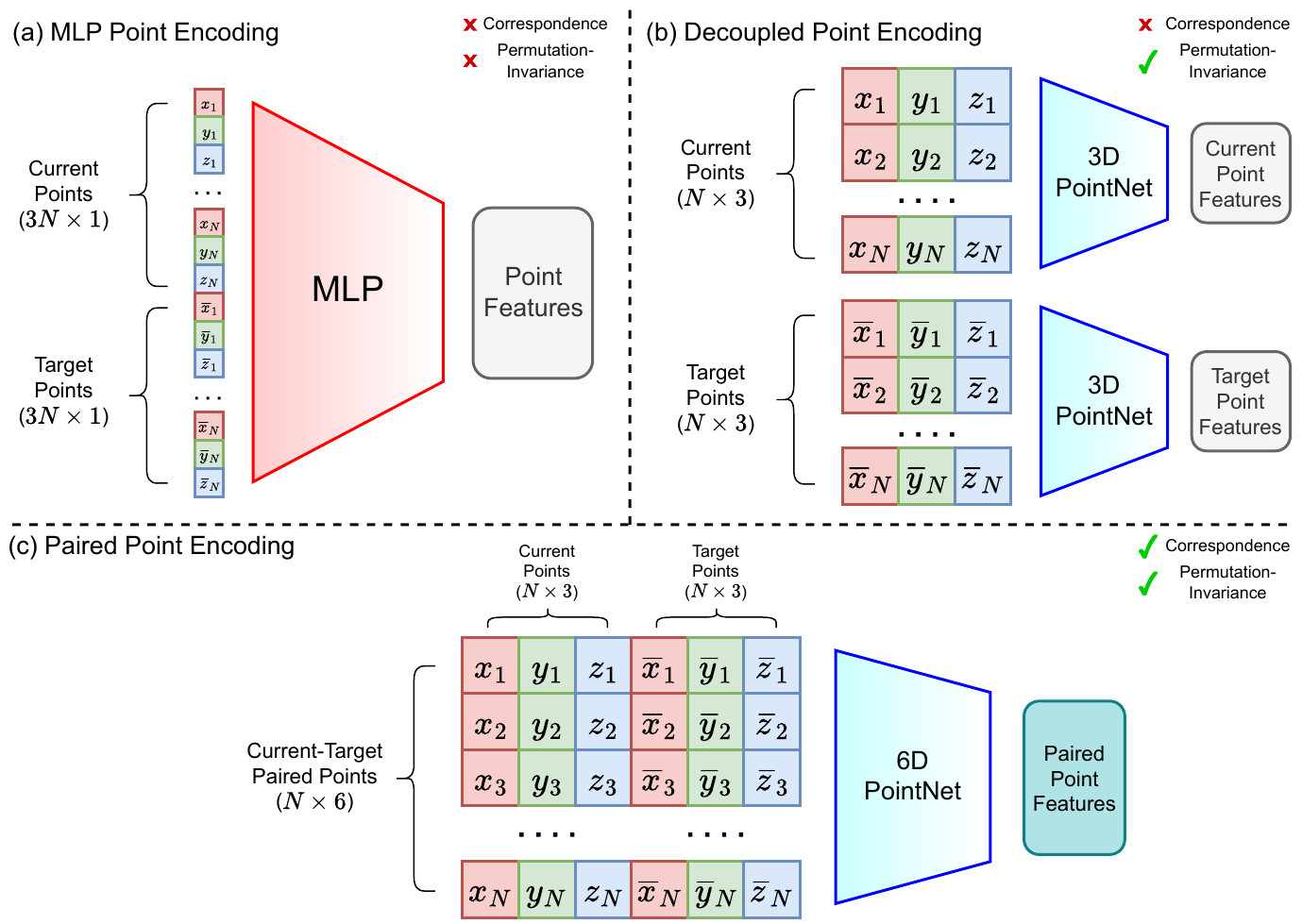}
  \caption{Comparison between our \representation~with other representations. Our point features keep \textbf{correspondence} and \textbf{permutation-invariance}, boosting performance for policy learning.}
  \label{fig:representation}
\end{wrapfigure}

\paragraph{Why point tracks?} A key design choice for training the AP2AP policy is selecting a goal representation that can be robustly extracted at deployment time while remaining flexible. In this work, we represent objects using sparse object point tracks, which are geometry-aware and can be reliably obtained in the real world using point trackers. More importantly, point track representation doesn't assume \textit{known mesh} and goes beyond \textit{rigid 6D poses}.

A common approach is to encode current and target object points separately into two latent features and condition the policy on them~\cite{lyu2025dywa}.
However, such encodings discard \textbf{correspondence} between current and target object points, which is critical for differentiating object poses.
To address this limitation, we propose \textbf{\representation}, a representation that explicitly preserves correspondence between the current and target object points. As illustrated in Fig.~\ref{fig:representation}, given the current object points $\{\bm{p}^i_t\}_{i=1}^N$ and the target object points $\{\bm{\bar{p}}^i_t\}_{i=1}^N$ at timestep $t$, we construct paired points $\{\bm{q}^i_t\}_{i=1}^N$ by concatenating each pair of corresponding points.
Each pair point is therefore 6-dimensional, consisting of a 3D current object point and its 3D target counterpart. For point index $i$ at timestep $t$, the paired point is defined as $\bm{q}_t^i = \begin{bmatrix} \bm{p}^i_t \\ \bm{\bar{p}}^i_t \end{bmatrix} \in \mathbb{R}^6$.

These paired points $\{\bm{q}^i_t\}_{i=1}^N \in \mathbb{R}^{N\times 6}$ are then fed into a PointNet-style encoder~\cite{qi2017pointnet, qi2017pointnet++}, which consists of shared MLP layers and mean-max mixed pooling to encode them into paired point features. In this way, we keep both \textbf{correspondence} and \textbf{permutation-invariance} of these points.
Building on the \representation~as goal representation, we now describe how it is used to train the AP2AP policy in simulation via a teacher–student learning framework. 

\subsection{Teacher-Student Policy Learning}
\label{sec:teacher-student}


\paragraph{\textbf{RL Teacher Policy Learning}}

We first train a teacher policy using PPO~\cite{schulman2017proximalppo} with privileged states and fully observed object geometry in simulation. As shown in Fig.~\ref{fig:method}, the state $s_t$ consists of robot proprioception (joint angles and velocities), the last action, and privileged information (e.g., joint torques, fingertip-to-object distances, etc.). Following Sec.~\ref{sec:representation}, we compose current and target object points into paired points as the goal representation $g_t$ and encode them using a lightweight PointNet~\cite{qi2017pointnet} to preserve both correspondence and permutation invariance. The resulting feature is concatenated with the state components and provided as input to the PPO actor and critic networks.


For reward shaping, instead of directly using the 6D pose (object position + rotation), our reward function leverages object points for a smoother reward landscape~\cite{zhao2025resmimic}. These rewards encourage the current object points to closely match the target object points, while promoting hand-object affinity and discouraging exaggerated motions:
\begin{equation}
    r = r_{\mathrm{goal}} + r_{\mathrm{f,o}} + r_{\mathrm{h,o}} + r_{\mathrm{bonus}} + r_{\mathrm{curl}} + r_{\mathrm{table}} + r_{\mathrm{action}}
\end{equation}
where $r_{\mathrm{goal}}$, $r_{\mathrm{f,o}}$, $r_{\mathrm{h,o}}$, $r_{\mathrm{bonus}}$, $r_{\mathrm{curl}}$, $r_{\mathrm{table}}$, and $r_{\mathrm{action}}$ represent rewards for current-target \textbf{point distances}, finger-object distances, hand-object distance, success bonus, finger curl, table collision penalty, and action penalty, respectively. More details on curriculum design and reward shaping are provided in Sec.~\ref{sec:app-curriculum} and Sec.~\ref{sec:app-reward}.


\paragraph{\textbf{Student Dynamics Action Model Learning}}

After training the teacher policy, we distill it into a student policy under partial observability using DAgger~\cite{ross2011reductiondagger}. We introduce a \textbf{transformer-based dynamics action model} that jointly learns action prediction and robot joint dynamics. This joint formulation improves action learning~\cite{kuang2024stopnet, lyu2025dywa, zhu2025unified, cen2025worldvla} and supports safer and more controllable deployment, particularly for high-DoF and highly dynamic hand–arm systems.

As illustrated in Fig.~\ref{fig:method}, the student policy takes as input robot proprioception (joint angles and velocities), the last action, and masked paired points. These inputs are first tokenized using MLPs and a PointNet-style encoder~\cite{qi2017pointnet, qi2017pointnet++} with mean–max mixed pooling, and then processed by self-attention layers~\cite{vaswani2017attention}, outputting robot action $a_t$ and next-state joint angle $\bm{\theta}_{t+1}$ and velocity $\dot{\bm{\theta}}_{t+1}$.

The dynamics action model is trained with a combination of a DAgger behavior cloning loss and a dynamics modeling loss:

\begin{equation}
\begin{aligned}
\mathcal{L}
  &= \mathcal{L}_{\mathrm{BC}} + \mathcal{L}_{\mathrm{DM}}
  = \lVert a_t^{\mathrm{stu}} - a_t^{\mathrm{tea}}\rVert_1 
        + \lVert
        \begin{bmatrix}
        \hat{\bm{\theta}}_{t+1} - \bm{\theta}_{t+1} \\ 
        \hat{\dot{\bm{\theta}}}_{t+1} - \dot{\bm{\theta}}_{t+1}
        \end{bmatrix}
        \rVert_1
\end{aligned}
\end{equation}
where $a_t^{\mathrm{stu}}$ and $a_t^{\mathrm{tea}}$ denote the student and teacher actions, respectively, and $\hat{\bm{\theta}}_{t+1}$, $\hat{\dot{\bm{\theta}}}_{t+1}$ and $\bm{\theta}_{t+1}$, $\dot{\bm{\theta}}_{t+1}$ denote the predicted and ground-truth next-step joint angles and velocities.

To emulate object point occlusions caused by the hand in real-world settings and improve robustness to monocular viewpoints, we introduce a \textbf{random plane-height masking strategy}.
Specifically, we sample a random plane through the object center and mask out points on one side of the plane, followed by sampling a random height that masks the majority of points above it and a minority below it. Target points are masked accordingly based on correspondence. This strategy enables the student policy to generalize across diverse camera viewpoints under partial observation.
More details can be found in Sec.~\ref{sec:app-masking}.

\subsection{Policy Deployment via Video Generation Planners}
\label{sec:whole-deployment}
To leverage the AP2AP policy in the real world, we condition it on desired 3D point tracks -- a sequence of target object points that specifies the desired object configuration over time. Such point tracks can be obtained from diverse sources, including video generation models or one-shot human demonstrations. In this section, we describe how we extract point tracks from generated videos and how they are used to drive closed-loop policy execution.

\subsubsection{From Generated Video to Object-Centric Point Tracks}
\label{sec:video-to-4d}
Large-scale video generation models provide a rich source of object motion and manipulation information, as they are trained on vast collections of Internet videos~\cite{mei2026video, patel2025robotic, li2025novaflow, chen2025large}. To use these videos for real-world robot deployment, we lift the videos into object-centric point tracks, which define a target trajectory that can directly condition the AP2AP policy.

Formally, given a language instruction $l$ and an initial RGBD observation $\{I_0, D_0\}$, we first generate a sequence of future RGB frames $\{I_t\}_{t=1}^T$ using an off-the-shelf video generation model. Using the initial object segmentation mask and the generated frames, we first perform \textbf{2D point tracking} to obtain object 2D point tracks $\{\bm{\bar{u}}_t^i\}_{t=1,i=1}^{T,N} \in \mathbb{R}^{T \times N \times 2}$.
Then we lift the 2D point tracks to 3D point tracks $\{\bm{\bar{p}}_t^i\}_{t=1,i=1}^{T,N} \in \mathbb{R}^{T \times N \times 3}$ using \textbf{relative depth estimation} with depth calibration using initial depth observation $D_0$.


The generated 3D point tracks provide a structured, object-centric representation of the desired pose over time, which can be directly used as a plan for the AP2AP point track policy, guiding the robot to perform pose-to-pose manipulation in the real world. Compared to prior approaches that rely on fully metric depth estimation and explicit calibration~\cite{patel2025robotic, li2025novaflow}, calibrating relative depth produces smoother and more stable metric depths, resulting in cleaner point tracks, as shown in Fig.~\ref{fig:depth}. Additional implementation details are provided in Sec.~\ref{sec:app-videogen}.

\subsubsection{Closed-Loop Perception and Control}
With the target point tracks defined, we now describe how the AP2AP policy executes it in a closed loop in the real world.
At the start of execution, the first set of target points is assigned to the policy. The initial tracked 2D points from the generated video are also provided for an online point tracker~\cite{karaev2025cotracker3}, which is used to track the object 2D points from the image observation in real time, which are then back-projected to 3D using the RGBD camera.

At each timestep, the current object 3D points and target object 3D points are composed into paired points and provided to the student policy along with robot proprioception and the last action. The policy then outputs actions for robot control. To determine when to advance to the next set of target points, we compute the average distance between corresponding visible points at each timestep:


\begin{equation}
    d_t = \frac{1}{N^\prime} \sum_{i=1}^{N^\prime} \lVert \bm{p}_t^i - \bm{\bar{p}}_t^i \rVert_2
\end{equation}

where $N^\prime$ is the number of visible points. When $d_t$ falls below a threshold, the target points are updated to the next ones in the goal 3D point tracks. This process repeats in a closed loop until the final target points $\{\bm{\bar{p}}_T^i\}_{i=1}^N$ are reached.


\begin{table*}[t]
\centering
\small
\setlength{\tabcolsep}{6pt}
\scalebox{0.75}{
\begin{tabular}{lccccccccccccccc}
\toprule
\multirow{2}{*}{\textbf{Method}} & \multirow{2}{*}{\textbf{CL}}
& \multicolumn{2}{c}{\textbf{\textit{Apple2Plate}}}
& \multicolumn{2}{c}{\textbf{\textit{Pour}}}
& \multicolumn{2}{c}{\textbf{\textit{Hammer}}}
& \multicolumn{2}{c}{\textbf{\textit{StackCup}}}
& \multicolumn{2}{c}{\textbf{\textit{RotateBox}}}
& \multicolumn{2}{c}{\textbf{\textit{Sponge2Bowl}}}
& \multicolumn{2}{c}{\textbf{Avg.}} \\
\cmidrule(lr){3-4}\cmidrule(lr){5-6}\cmidrule(lr){7-8}\cmidrule(lr){9-10}\cmidrule(lr){11-12}\cmidrule(lr){13-14}\cmidrule(lr){15-16}
&
& \textbf{SR} & \textbf{TP}
& \textbf{SR} & \textbf{TP}
& \textbf{SR} & \textbf{TP}
& \textbf{SR} & \textbf{TP}
& \textbf{SR} & \textbf{TP}
& \textbf{SR} & \textbf{TP}
& \textbf{SR} & \textbf{TP} \\
\midrule
NovaFlow~\cite{li2025novaflow}
& \nomark
& 0.44 & 0.57 
& 0.66 & 0.76 
& 0.12 & 0.16 
& 0.42 & 0.59 
& 0.25 & 0.32 
& 0.18 & 0.29 
& 0.345 & 0.448 \\
LVP~\cite{chen2025large}
& \nomark
& 0.26 & 0.64 
& 0.39 & 0.42 
& 0.00 & 0.00 
& 0.10 & 0.61 
& 0.00 & 0.16 
& 0.11 & 0.41 
& 0.143 & 0.373 \\
NovaFlow-CL~\cite{li2025novaflow}
& \yesmark
& 0.58 & 0.76
& 0.73 & 0.85
& 0.27 & 0.30
& 0.44 & 0.73
& 0.33 & 0.60
& 0.27 & 0.41 
& 0.437 & 0.608 \\
\method{} \textbf{(Ours)}
& \yesmark
& \textbf{0.86} & \textbf{0.91}
& \textbf{0.86} & \textbf{0.92}
& \textbf{0.28} & \textbf{0.31}
& \textbf{0.60} & \textbf{0.82}
& \textbf{0.44} & \textbf{0.69}
& \textbf{0.56} & \textbf{0.62}
& \textbf{0.600} & \textbf{0.712} \\
\bottomrule
\end{tabular}
}
\caption{Quantitative results measured by Success Rate \textbf{(SR)} and Task Progress \textbf{(TP)} in simulation. \textbf{CL} stands for closed-loop. All tasks are \textbf{zero-shot}. We adapt NovaFlow and NovaFlow-CL to dexterous hands by using our method for first-stage dexterous grasping.}
\label{tab:sim_comparison}
\vspace{-10pt}
\end{table*}

\section{Experiments}

\subsection{Simulation Experiments}

\paragraph{Experimental Setup}

We evaluate our method and baselines on six dexterous manipulation tasks, namely \textit{Apple2Plate}, \textit{Pour}, \textit{Hammer}, \textit{StackCup}, \textit{RotateBox}, and \textit{Sponge2Bowl}. All tasks are completely \textbf{unseen and zero-shot}. We compare against NovaFlow~\cite{li2025novaflow}, Large Video Planner (LVP)~\cite{chen2025large} and a closed-loop version of NovaFlow (NovaFlow-CL). Details on task specifications and baseline implementation can be found in Sec.~\ref{sec:app-task-specifications} and Sec.~\ref{sec:app-baseline}.

In the simulation, we manually pre-define several waypoints as goal poses and compute the target points with random Gaussian noise for policy condition. The goal pose will update once the current goal pose is achieved. To measure performance, we use Success Rate \textbf{(SR)} and Task Progress \textbf{(TP)} as quantitative metrics. Success is determined by whether the last waypoint is achieved or not, and Task Progress is defined as the average number of waypoints achieved. For each trial, we initialize the object with a random pose, and we conduct 100 trials for each task.

\paragraph{\textbf{Results and Analysis}}

We report our simulation results in Table.~\ref{tab:sim_comparison}. As shown in the results, our method outperforms all baselines by a large margin, demonstrating the effectiveness and superiority of our method.


Compared with open-loop NovaFlow and LVP, our method shows a clear advantage in closed-loop reactivity. Closed-loop feedback greatly improves performance since the object may move around in hand. In such cases \method{} can dynamically adjust arm and finger movements to keep the object on track, while motion planning or retargeting are highly insufficient under such dynamic scenarios.
For LVP, which leverages video generation model to generate a video given target task, followed by retargeting to the robot hand, it suffers from the embodiment gap and a lack of closed-loop feedback, and therefore performs the worst.

Further comparing our method with NovaFlow-CL, we also surpass it, achieving a significant performance gain ($+16.3\%$ in SR and $+10.4\%$ in TP). This demonstrates that our policy, trained on AP2AP, exhibits strong generalizability to unseen tasks, objects, and trajectories, despite both methods utilizing closed-loop feedback. We attribute this performance gain to the fact that during the RL and DAgger process, the policy is able to traverse sufficient states so that it can generalize to unseen scenarios. NovaFlow-CL fails mainly because of insufficient hand reactivity to high dynamics that causes object falling, limited motion planning solutions that severely occlude object points, and pose estimation errors due to point noises, especially when the number of visible points is low. This further highlights our method's robustness and adaptivity to out-of-distribution scenarios and effectiveness in generalizing across objects and scenes.

\subsection{Ablation Studies}
\label{sec:ablation}

\begin{wrapfigure}{r}{0.4\textwidth}
\centering
\includegraphics[width=1.0\linewidth]{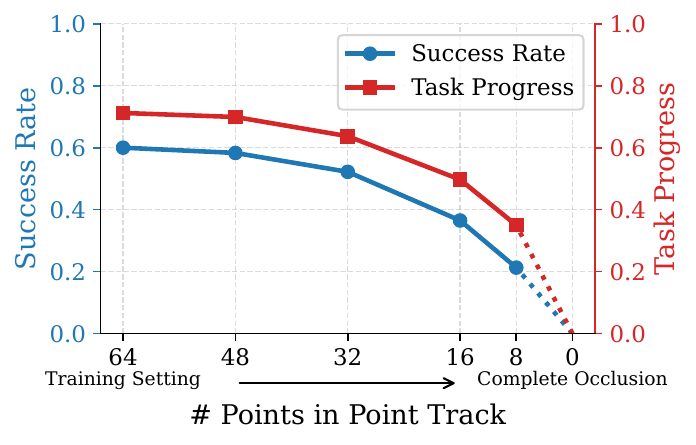}
  \caption{Our point track policy is robust to different degrees of point loss.}
  \label{fig:point_track_num}
\end{wrapfigure}

To further analyze our framework design, we conduct ablation studies on various components to highlight the importance of each element in our method. Ablation results are shown in Table~\ref{tab:ablation_stu}, Fig.~\ref{fig:point_track_num}, and Fig.~\ref{fig:ablation_tea}, showing that our method surpasses all ablations, proving the effectiveness of all the modules.

\begin{wraptable}{r}{0.4\textwidth}
\centering
\small
\setlength{\tabcolsep}{6pt}
\scalebox{0.9}{
\begin{tabular}{lcc}
\toprule
\textbf{Method} & \textbf{SR} & \textbf{TP} \\
\midrule
MLP Point Encoding
& 0.057 & 0.172 \\
Decoupled Point Encoding
& 0.203 & 0.363 \\
w/o Self-Attention
& 0.490 & 0.675 \\
w/o Dynamics Modeling
& 0.570 & 0.683 \\
\method{} \textbf{(Ours)}
& \textbf{0.600} & \textbf{0.712} \\
\bottomrule
\end{tabular}
}
\caption{Ablation studies on the student policy. \textbf{SR} and \textbf{TP} are task-averaged.}
\label{tab:ablation_stu}
\end{wraptable}

\paragraph{\textbf{Importance of~\representation}}
\label{sec:ablation-representation}

First, we discuss whether using our proposed \representation~can improve policy performance. 
We conduct experiments on our student policy, as shown in Table~\ref{tab:ablation_stu}. Our finding suggests that using MLP to encode the points leads to severe performance degradation, making SR reduce to $5.7\%$. We also find out that using separate PointNet encoders to encode current and target object points individually will also severely lower the performance since the policy loses the correspondence between the two sets of points. This highlights our representation's advantage to effectively encode the object geometry information and point correspondence. We also ablate the importance of \representation{} on the teacher policy in Sec.~\ref{sec:app-ablation} and Fig.~\ref{fig:ablation_tea}.


\paragraph{\textbf{Ablation on Policy Architecture}}

Furthermore, we also ablate different neural network architecture design choices for our student policy.
As shown in Table~\ref{tab:ablation_stu}, compared with \textit{w/o Self-Attention} that naïvely concatenates features and uses MLP to decode actions, self-attention layers can attend to different tokens from proprioception and paired points, which better captures relations from different input components. We also find that integrating dynamics modeling can improve performance, which correlates well with the synergistic effects of policy learning and dynamics modeling.

\paragraph{\textbf{Robustness to Point Number in the Point Track}}

We also ablate how our policy performs under various degrees of point loss in the point track, as points could be invisible under heavy occlusions. As shown in Fig.~\ref{fig:point_track_num}, compared to the full training setting (64 points), reducing the point number to 48 and 32 only results in very limited performance degradation. Moreover, our policy could still function even if there are less than 10 points left, demonstrating our method's robustness under extreme occlusion settings. 
In the real world there are often more than 64 points visible (100+) on the object so these extreme scenarios would rarely happen.

\subsection{Real-World Experiments}

We deploy our policy to the real world as the procedure detailed in Sec.~\ref{sec:whole-deployment}. Fig.~\ref{fig:teaser} and Fig.~\ref{fig:gallery} show a gallery of tasks our method can do in a completely zero-shot manner. It shows the strong \textbf{zero-shot} generalization capabilities of our method on diverse tasks, objects, scenes, and perturbations. For more deployment results, check out Fig.~\ref{fig:gallery} and our supplementary video.

Specifically, on four tasks, namely \textit{LiftToy}, \textit{Broccoli2Plate}, \textit{Meat2Bowl}, and \textit{Pour},
we compare our method against various baselines, including NovaFlow-CL~\cite{li2025novaflow}, LVP~\cite{chen2025large}, as well as $\pi_{0.5}$-DROID to further show the inadequacy of VLA models on these open-world tasks. Note that $\pi_{0.5}$-DROID uses parallel grippers which are much easier for these tasks.
All tasks and objects are \textbf{unseen and zero-shot} and there are \textbf{no real robot demonstrations} for any task.




\begin{wraptable}{r}{0.6\textwidth}
\centering
\small
\setlength{\tabcolsep}{6pt}
\scalebox{0.8}{
\begin{tabular}{lccccc}
\toprule
\textbf{Method}
& \textbf{\textit{LiftToy}}
& \textbf{\textit{Broccoli2Plate}}
& \textbf{\textit{Meat2Bowl}}
& \textbf{\textit{Pour}}
& \textbf{Total} \\
\midrule
LVP~\cite{chen2025large}
& 9/20
& 6/20
& 7/20
& 0/20
& 22/80
\\
NovaFlow-CL~\cite{li2025novaflow}
& 10/20
& 7/20
& 7/20
& 0/20
& 24/80
\\
$\pi_{0.5}$-DROID*~\cite{intelligence2025pi05}
& 5/20
& 6/20
& 9/20
& 0/20
& 20/80
\\
\method{} \textbf{(Ours)}
& \textbf{14/20}
& \textbf{10/20}
& \textbf{13/20}
& \textbf{8/20}
& \textbf{45/80}
\\
\bottomrule
\end{tabular}
}
\caption{Quantitative results measured by Success Rate \textbf{(SR)} in the real world. All tasks are \textbf{zero-shot}, all objects are \textbf{unseen} and there are \textbf{no} real robot demonstrations. * denotes using parallel grippers.}
\label{tab:real_comparison}
\end{wraptable}

As shown in Table~\ref{tab:real_comparison}, our method achieves a 87.5\% relative performance gain in \textbf{SR} compared to the best-performing baseline (NovaFlow-CL), demonstrating our method's superiority against the motion planning, VLAs, and plain retargeting. Our superiority mainly comes from the generalization from video models and our policy's closed-loop reactivity for both arm and hand, as well as robust action prediction under noisy point input, as illustrated in Fig.~\ref{fig:qual_real}.

\subsection{Failure Analysis}

\begin{wrapfigure}{r}{0.45\textwidth}
\centering
\includegraphics[width=1.0\linewidth]{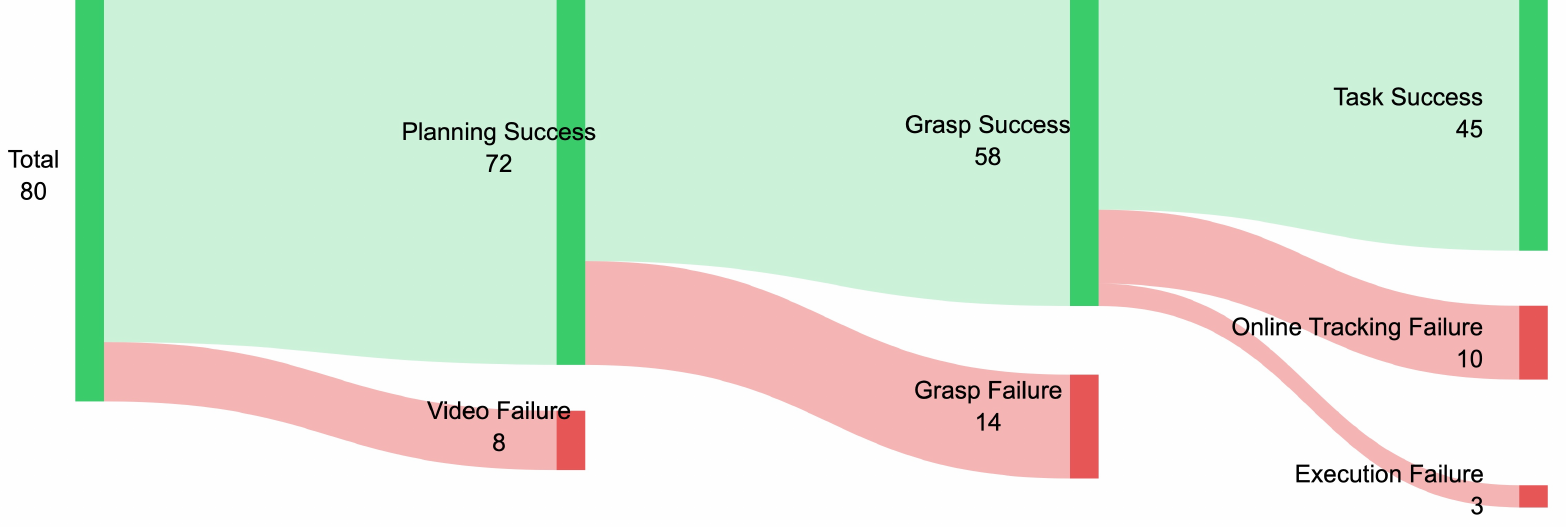}
  \caption{Failure analysis of our method.}
  \label{fig:failure}
\end{wrapfigure}


To further investigate the failure modes of our method, we conduct a failure analysis as shown in Fig.~\ref{fig:failure}.
First, video generation hallucinations serve as the \textbf{only} point of failure of our high-level planning pipeline. This suggests that the offline tracking and depth estimation components are already sufficiently robust, contrary to the common concern that imperfections in these foundation models would lead to compounding errors throughout the pipeline.
On the robot execution side, sometimes the policy also tends to push the object to form a firm grasp, but this might pose extra forces that in turn knocks over the object, causing grasp failures. 
Finally, the real-time online tracker would lose track of the object (causing 0 point in the point track) when there are significant object movements or similar nearby textures.

\begin{figure}[t]
\centering
\includegraphics[width=1.0\linewidth]{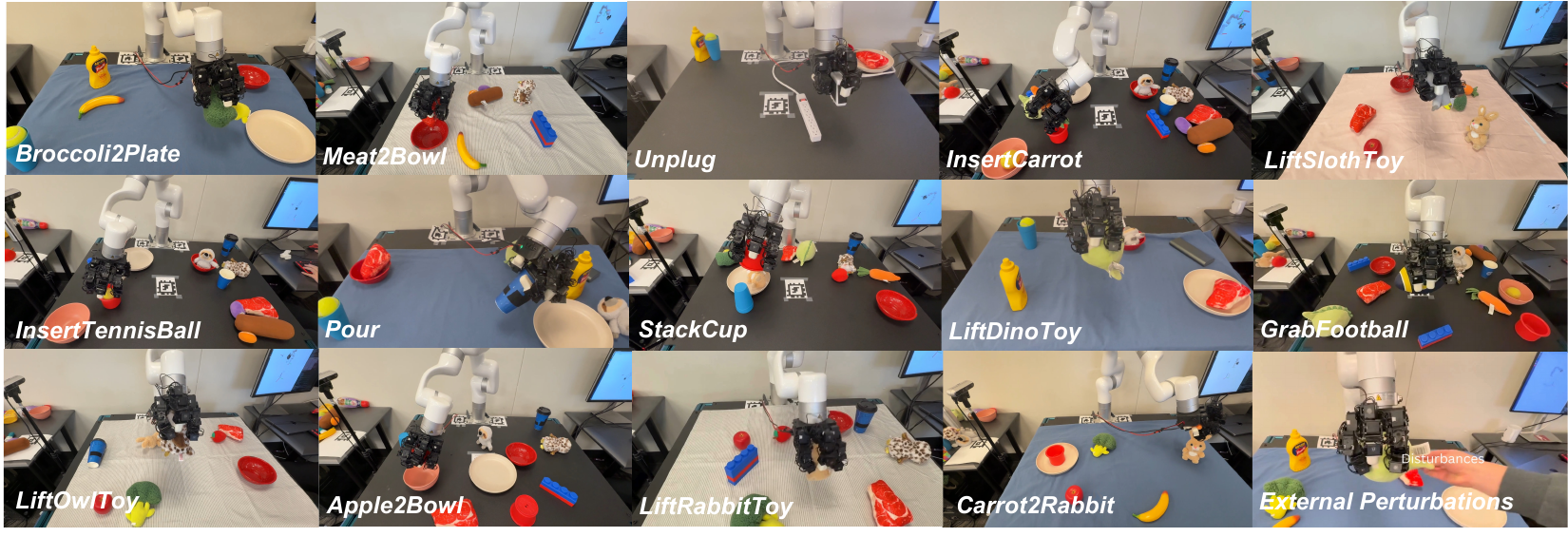}
  \caption{Our \method{} can achieve a wide variety of open-world tasks in a \textbf{zero-shot} manner, generalizing across tasks, objects, scenes, and external perturbations.}
  \label{fig:gallery}
\end{figure}
\section{Conclusions, Limitations, and Future Works}

\textbf{Conclusions.} In this work, we propose \method, a framework for generalizable dexterous manipulation via object-centric point tracks and task-agnostic sim-to-real learning.
At the core of \method{} is to decouple recognition and control by leveraging video generation and 4D reconstruction to generate object-centric point tracks as high-level planning, and training a task-agnostic sim-to-real policy for low-level control.
We further propose a novel \representation~representation and a transformer-based dynamics action model to enhance 3D goal-conditioned policy learning. Extensive experiments in simulation and the real world verify the effectiveness of our framework, and show our remarkable generalization to unseen tasks, objects, and scenes. 

\textbf{Limitations and Future Works.} Despite compelling results, our work has certain limitations that can be further improved in future works. 
First, in our work we didn't incorporate human grasp priors from HOI datasets and Internet videos due to the lack of amount and diversity of clean mocap sequences and the large embodiment gap between human hands and the LEAP hand, which is large in size and only has four thick fingers with limited dexterity. It's promising to leverage these abundant hand-related data sources along with thinner and more human-like dexterous hands to unlock more functional behaviors and finer fingertip control. 
Moreover, our AP2AP formulation is currently limited to single-object manipulation. Extending it to objects with more complicated geometries, such as articulated or deformable objects, would be a promising direction. 
Finally, our closed loop is reactive to the \emph{manipulated} object: online tracking lets the policy recover from in-hand slip and external perturbations. It does not observe the target or scene obstacles. The natural remedy within our framework is high-level replanning, i.e., regenerating the video plan and point tracks from the current observation, which is compatible with our interface but not implemented here, and would be a natural extension.

\section*{Acknowledgments}
We express our sincere gratitude to Jiashun Wang, Jialiang Zhang, and Andrew Wang for fruitful discussions, to Kenny Shaw for hardware support, and to Himangi Mittal, Minsik Jeon, Yehonathan Litman, Qitao Zhao, Zihan Wang, Hanzhe Hu, and Lucas Wu for presentation feedback.
This work was supported in part by gifts from Google and CISCO, NSF Award IIS-2442282, DARPA SAFROn award HR0011-25-3-0203, NSF Career award, an Amazon robotics award, and AFOSR Grant FA9550-23-1-0257.


\clearpage
\bibliography{references}  

 \clearpage
 \appendix
 \appendix
\section{Videos and Visualizations}

Videos and visualization results can be found in our supplementary video and project page: \texttt{\href{https://dex4d.github.io/}{\textcolor{cmured}{https://dex4d.github.io/}}}. We use Viser~\cite{yi2025viser} for all the visualizations. We thank the authors for their great work.

\section{Task Specifications}
\label{sec:app-task-specifications}

\begin{figure*}[t]
\centering
\includegraphics[width=1.0\linewidth]{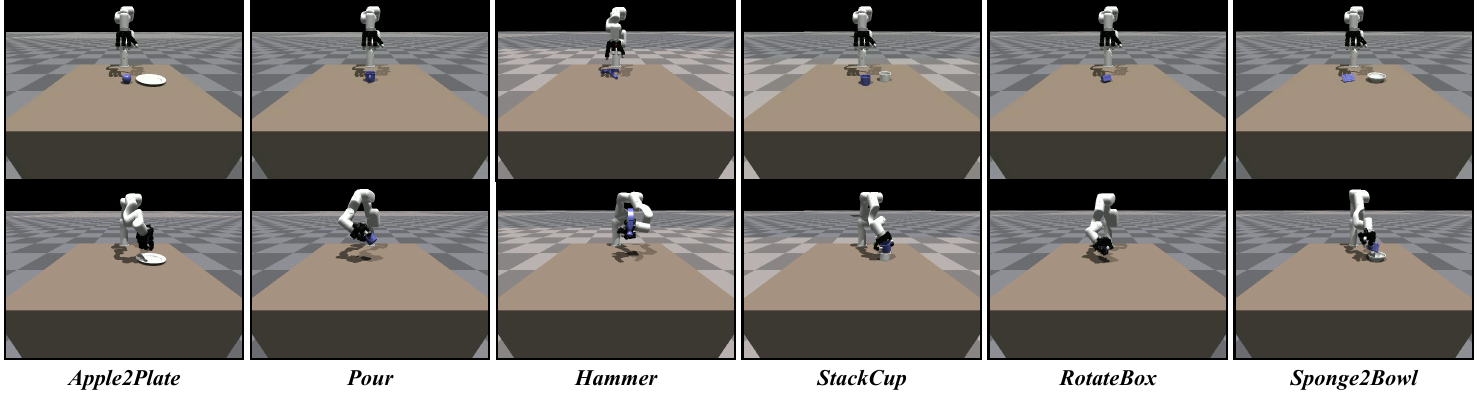}
  \caption{Overview of simulated tasks. Two frames are shown in each column for each task.}
  \label{fig:sim_tasks}
\end{figure*}
\begin{figure}[t]
\centering
\includegraphics[width=1.0\linewidth]{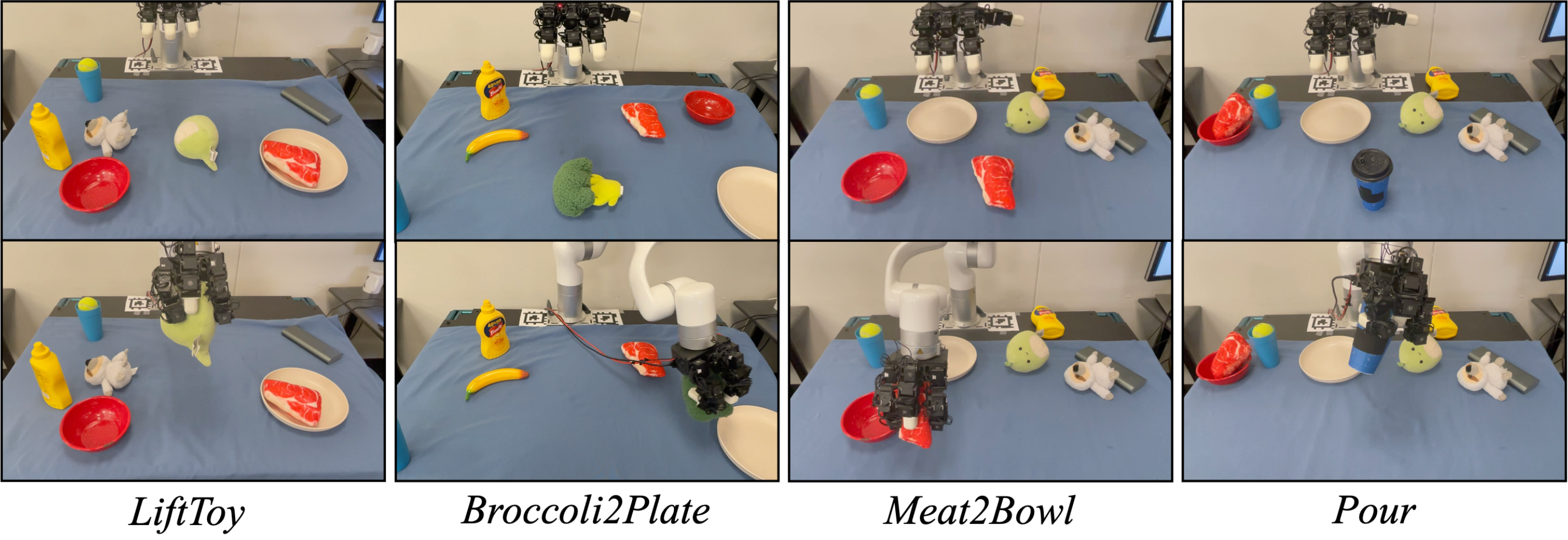}
  \caption{Overview of our real-world benchmarking suites. Two frames are shown in each column for each task.}
  \label{fig:tasks}
\end{figure}
\begin{figure*}[t]
\centering
\includegraphics[width=1.0\linewidth]{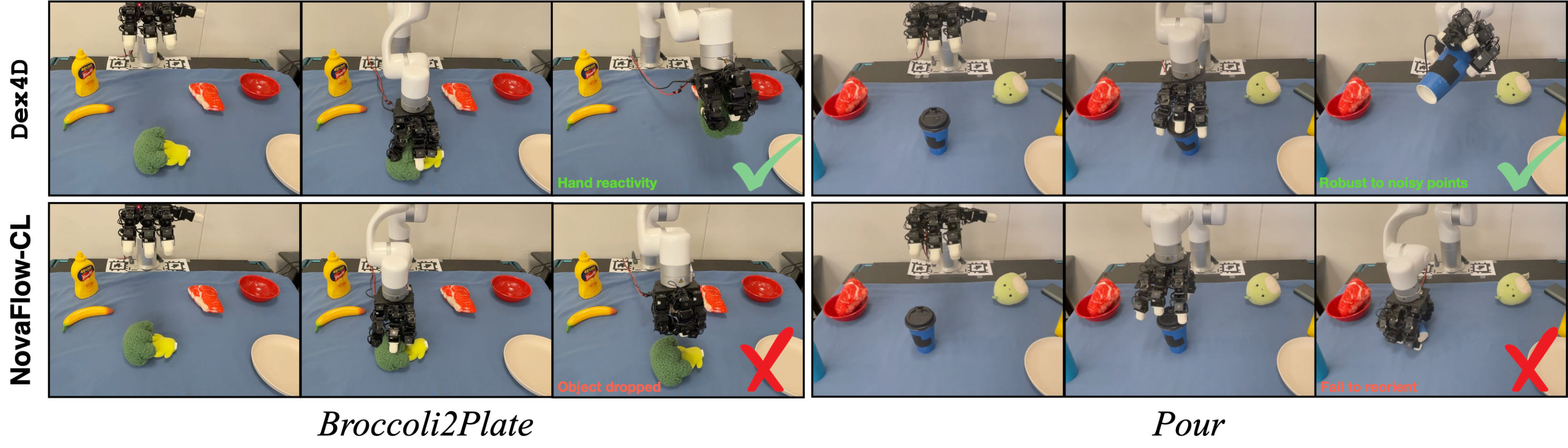}
  \caption{Qualitative comparison between our method and the NovaFlow-CL baseline. The baseline suffers from object dropping and inaccurate post-grasping movement due to the \textbf{lack of hand feedback} and \textbf{vulnerability to few and noisy visible points}, while our method performs robustly.}
  \label{fig:qual_real}
\end{figure*}

In simulation, we evaluate our method and baselines on six dexterous manipulation tasks, namely \textit{Apple2Plate}, \textit{Pour}, \textit{Hammer}, \textit{StackCup}, \textit{RotateBox}, and \textit{Sponge2Bowl}. We illustrate these tasks in Fig.~\ref{fig:sim_tasks}. Their task objectives are:

\begin{itemize}
    \item \textit{Apple2Plate}: Grasp an apple on the table and put it on the plate.
    \item \textit{Pour}: Grasp the mug on the table and tilt to pour.
    \item \textit{Hammer}: Grasp the hammer on the table and strike forward.
    \item \textit{StackCup}: Pick up the cup and stack it onto another cup on the table.
    \item \textit{RotateBox}: Pick up the Foam Box on the table and rotate it horizontally 90 degrees in the air.
    \item \textit{Sponge2Bowl}: Grasp a thin piece of sponge on the table and put it into the bowl.
\end{itemize}

In the real world, we evaluate on four tasks, namely \textit{LiftToy}, \textit{Broccoli2Plate}, \textit{Meat2Bowl}, and \textit{Pour}, as shown in Fig.~\ref{fig:tasks}. Their task objectives are:

\begin{itemize}
    \item \textit{LiftToy}: Grasp a toy on the table and lift it to a certain pose in the air.
    \item \textit{Broccoli2Plate}: Pick up the broccoli on the table and put it on the plate.
    \item \textit{Meat2Bowl}: Pick up the meat on the table and put it into the bowl.
    \item \textit{Pour}: Grasp the coffee cup on the table and tilt to pour.
\end{itemize}

\section{Implementation Details}

In our work, we use a 22-DoF dexterous hand-arm system that comprises a 6-DoF xArm6 robot arm and a 16-DoF LEAP hand~\cite{shaw2023leap}. We use a single-view RealSense D435 camera for RGBD sensing and Apriltags~\cite{olson2011apriltag} for calibration.

For the action space, the policy outputs a 22-dimensional action, which contains 6-DoF arm delta joint angles and 16-DoF LEAP hand absolute joint angles. We found that this configuration is more robust in RL training without compromising on full hand dexterity~\cite{ciocarlie2007dexterouseigengrasp, agarwal2023dexterousfunctionalgrasping, lum2024dextrahg}. The action is then converted to the 22-DoF joint target used for motor control. The rationale behind the action space design is that arm movements are large, so delta commands bound per-step motion and stabilize RL exploration, whereas finger targets require absolute precision and are better predicted directly; this split follows prior dexterous sim-to-real works~\cite{huang2023dynamic}.

For desired point tracks acquisition, we use Wan2.6~\cite{wan2025wan} for video generation, Video Depth Anything~\cite{chen2025video} for relative video depth estimation, SAM2~\cite{ravi2024sam2} for initial frame segmentation, and CoTracker3~\cite{karaev2025cotracker3} for 2D point tracking. All these components are modular, reusable, and replaceable, which can be updated and swapped to new state-of-the-art models.

We train teacher and student policies using the Isaac Gym simulator~\cite{makoviychuk2021isaac}. The AP2AP policy is trained entirely in simulation using \textbf{3,200 objects} from UniDexGrasp~\cite{xu2023unidexgrasp} under diverse pose configurations and extensive domain randomization.

For point numbers, we sample 128 points for teacher training, and 64 points for student training and inference. For RL teacher training, we use symmetric PPO~\cite{schulman2017proximalppo} to learn 15k steps for the first stage, 10k steps for the second stage, and another 25k steps for the third stage to ensure convergence. For student policy learning, we select the best teacher checkpoint and learn 25k steps using DAgger~\cite{ross2011reductiondagger}. It takes 2-3 days for teacher training, and around 20 hours for student training, using a single NVIDIA RTX A6000 GPU.

For goal sampling, the ``nearby'' goal definition is that: the next goal is sampled within 10\,cm translation and 0.5\,rad rotation of the achieved pose.

\section{Baseline Details}
\label{sec:app-baseline}

\paragraph{NovaFlow~\cite{li2025novaflow}} NovaFlow is a method that leverages 3D actionable point tracks for robot manipulation. NovaFlow first extracts 3D action point tracks from generated videos, and then applies a grasp generator~\cite{murali2025graspgen} and performs open-loop motion planning based on transformation estimation between current and target object points using the Kabsch algorithm~\cite{kabsch1976solution}. Since NovaFlow is parallel gripper-only, for fair comparison, we adapt it to dexterous hands by applying our method for dexterous grasping and locking the fingers after lifting. Then we conduct pose estimation based on the current state and perform motion planning in an open-loop manner.

\paragraph{NovaFlow-CL~\cite{li2025novaflow}} Since NovaFlow is open-loop without feedback, which is crucial for high-dynamics tasks like dexterous manipulation, we also implemented a closed-loop version of NovaFlow (NovaFlow-CL) as a baseline. Specifically, we first apply our method for dexterous grasping, and then at each timestep, we estimate the transformation between current and target object points, and do real-time IK to move the robot arm accordingly.

\paragraph{Large Video Planner (LVP)~\cite{chen2025large}} LVP is a method that first uses video generation to generate hand-object interaction for one task, and then uses vanilla kinematic retargeting to obtain robot actions. Since the embodiment gap is so large for the LEAP hand that it can barely grasp the object with the retargeted hand pose, we first apply our method for grasping, and then execute the retargeted motion.

\paragraph{$\pi_{0.5}$-DROID~\cite{intelligence2025pi05}} A VLA baseline that is pretrained on vast amount of cross-embodiment data and then finetuned with DROID dataset. We use the same DROID (Franka + Robotiq parallel gripper) setting for policy inference.

\section{Additional Ablations}
\label{sec:app-ablation}

\begin{figure}[t]
\centering
\includegraphics[width=0.5\linewidth]{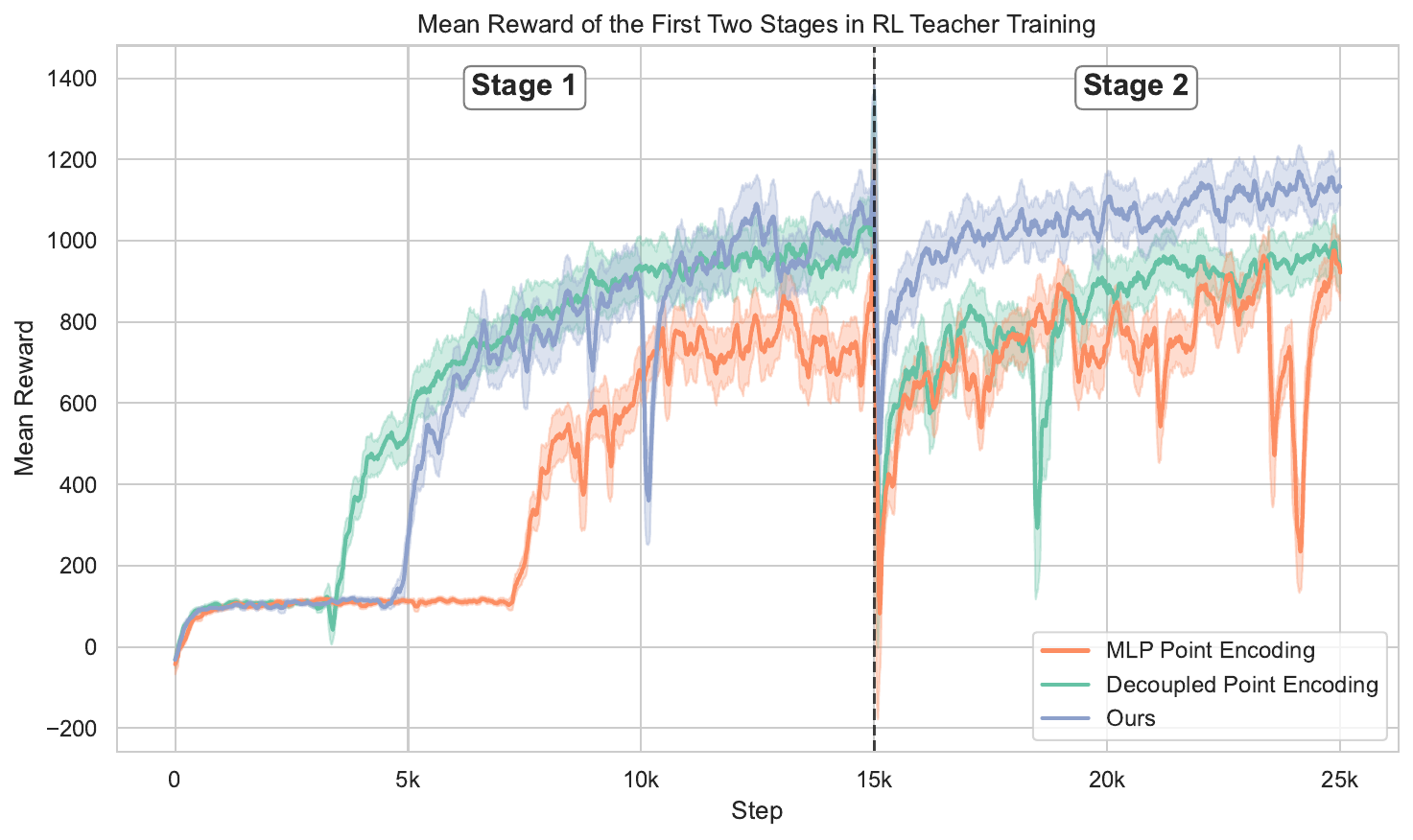}
  \caption{Mean reward curve of the first two stages of teacher training. Step 15k is the curriculum boundary. Our method outperforms both ablation variants.}
  \label{fig:ablation_tea}
\end{figure}

We also ablate with \textit{MLP Point Encoding} and \textit{Decoupled Point Encoding} in our RL \textbf{teacher} training to verify the effectiveness of our \representation~representation in RL training. As shown in Fig.~\ref{fig:ablation_tea}, our method outperforms both variants, demonstrating that our representation and framework can even boost the performance for visual RL, which is considered harder in prior works~\cite{kuang2025skillblender, lin2025sim, singh2025end}.

\section{Additional Result Analysis}

We also show the qualitative comparison in Fig.~\ref{fig:qual_real} and the supplementary video. As we can see, the baseline fails due to a couple of reasons. First, since the baseline is unaware of the hand and object grasping, the object would gradually fall off the hand during arm moving due to the lack of feedback. In contrast, our method learns to adjust or regrasp the object and then proceeds with the task. Moreover, in the real world the 3D point tracking poses large amounts of noise, including ones coming from inaccurate 2D tracking, noisy depth sensing, and latency, especially when the LEAP hand fingers severely occlude the object. Since the baseline method leverages the Kabsch algorithm~\cite{kabsch1976solution} to solve the 6D pose, it's prone to noisy observation especially when visible points are few. The Kabsch algorithm can hardly solve the correct rotation between noisy current and target object points under real-world scenarios when the object is occluded by the hand, as in the \textit{Pour} task, the baseline has a 0 success rate. In contrast, our method remains robust even if there are less than 10 visible points left. Finally, some failures of the baseline come from limited solutions of motion planning since we use a 6-DoF xArm6, which completely occludes the object from the camera.

Moreover, our method is robust to various generalization tests in the real world. As shown in Fig.~\ref{fig:teaser} and Fig.~\ref{fig:gallery}, although our policy is only trained on single-object scenarios purely in simulation, it generalizes well to unseen object types and poses, backgrounds, camera views, task trajectories, and external disturbances.

To reflect real-world per-task variance, besides absolute success rate as reported in Table~\ref{tab:real_comparison}, we also report per-task Wilson 95\% confidence intervals: \emph{LiftToy} 14/20=0.70 [0.48, 0.86], \emph{Broccoli2Plate} 10/20=0.50 [0.30, 0.70], \emph{Meat2Bowl} 13/20=0.65 [0.43, 0.82], \emph{Pour} 8/20=0.40 [0.22, 0.61], overall 45/80=0.56 [0.45, 0.67]. The intervals are wide at $n{=}20$.
The real-world experiment protocol is that, for each task we run 20 trials; the object is initialized at a position sampled over a 30\,cm$\times$30\,cm region with random rotations and surrounding clutter; the scene is fully reset between trials; and all methods are evaluated on the same set of initial configurations.

\section{Video Generation and Point Track Extraction}
\label{sec:app-videogen}

For video generation, we use Wan2.6~\cite{wan2025wan} with its online platform, and use its native prompt enhancement with Chinese prompts, which show better performance than English prompts~\cite{li2025novaflow}. We use Wan's first frame + language prompt conditioned generation mode and generate 5-second 30-FPS 720P videos.

\begin{figure}[t]
\centering
\includegraphics[width=1.0\linewidth]{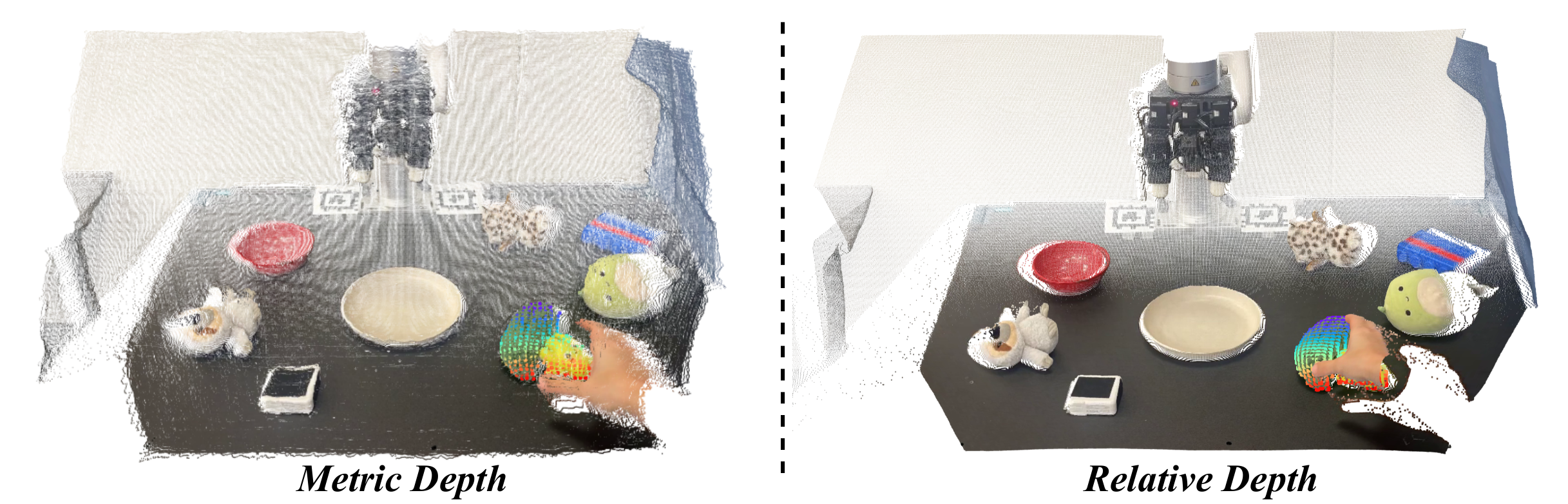}
  \caption{Comparison between metric depth estimation and relative depth estimation. Relative depth estimation yields smoother, more spatio-temporally consistent results and fewer floater points.}
  \label{fig:depth}
\end{figure}

For 3D point track extraction, we find that using relative depth estimation, rather than metric video depth estimation as in prior work~\cite {patel2022learning, li2025novaflow}, yields better results, with greater spatio-temporal consistency and fewer floaters, as shown in Fig.~\ref{fig:depth}.

\section{RL State Space}

The state space $\mathcal{S}$ for the RL teacher policy training includes: joint angles, joint velocities, the last action, joint torques, fingertip states (\textit{state} denotes 6D pose, linear and angular velocities, hereinafter the same), fingertip forces, hand state, object state, goal pose, 64-dimensional object point cloud feature encoded by a pretrained PointNet~\cite{qi2017pointnet, xu2023unidexgrasp}, and fingertip-to-object distance vectors.

\section{RL Curriculum Learning}
\label{sec:app-curriculum}

We detail our three-stage curriculum in Table~\ref{tab:curriculum}.

\begin{table}[t]
\centering
\small
\setlength{\tabcolsep}{6pt}
\scalebox{1.0}{
\begin{tabular}{lccc}
\toprule
& \textbf{Stage 1}
& \textbf{Stage 2}
& \textbf{Stage 3} 
\\
\midrule
Step
& 0-15k
& 15k-25k
& 25k-50k
\\
Object Category
& Bottle
& Bottle
& All
\\
\makecell[l]{Too Far Reset Threshold (m)} 
& 0.3
& $1\times 10^6$
& $1\times 10^6$
\\
Goal Reset Stable Ratio
& 0.1
& 0.2
& 0.2
\\
Arm Speed Scale
& 10
& 1.5
& 1.5
\\
Control Frequency (Hz)
& 30
& 30
& 5
\\
Step Size
& $3\times 10^{-4}$
& $3\times 10^{-4}$
& $3\times 10^{-5}$
\\
Initial Hand Position
& Low
& Low
& High
\\
\bottomrule
\end{tabular}
}
\caption{Curriculum settings of our three-stage RL teacher training.}
\label{tab:curriculum}
\end{table}

\section{Reward Function}
\label{sec:app-reward}

We detail our reward shaping in Table~\ref{tab:reward}. $p_j^{\mathrm{finger}}$, $p^{\mathrm{obj}}$, $p^{\mathrm{hand}}$, $h_j^{\mathrm{finger}}$ represent the 3D position of the finger $j$, the 3D position of the object, the 3D position of the hand palm and the height of the finger $j$, respectively.
And conditions $\mathrm{contact}$, $\mathrm{success}$, and $\mathrm{stay\_success}$ are defined as follows:

$$
\mathrm{contact}=
\begin{cases}
1, &
\begin{aligned}
\sum_{j=1}^4 d(p_{j}^{\mathrm{finger}},p^{\mathrm{obj}}) < 0.48\,\mathrm{m} \\
\text{and } d(p^{\mathrm{hand}},p^{\mathrm{obj}}) < 0.12\,\mathrm{m}
\end{aligned}
\\
0, & \text{otherwise}
\end{cases}
$$

$$
\mathrm{success} = 
\begin{cases}
    1, & d_t < 0.05\,\mathrm{m}
    \\
    0, & \text{otherwise}
\end{cases}
$$

$$
\mathrm{stay\_success}=
\begin{cases}
    1, & \mathrm{success\_time} \geq 0.5\,\mathrm{s}
    \\
    0, & \text{otherwise}
\end{cases}
$$

\begin{table}[t]
\centering
\small
\setlength{\tabcolsep}{6pt}
\scalebox{0.75}{
\begin{tabular}{lccc}
\toprule
\textbf{Term}
& \textbf{Expression}
& \textbf{Weight (Stage 1 \& 2)}
& \textbf{Weight (Stage 3)}
\\
\midrule
$r_{\mathrm{goal}}$
& $\mathbbm{1}_\mathrm{contact}(1.4-3d_t)$
& 1.0
& 1.0
\\
$r_{\mathrm{f,o}}$
& $-0.5\times \sum_{j=1}^4 d(p_{j}^{\mathrm{finger}},p^{\mathrm{obj}})$
& 1.0
& 0.0
\\
$r_{\mathrm{h,o}}$
& $-0.5\times d(p^{\mathrm{hand}},p^{\mathrm{obj}})$
& 1.0
& 0.0
\\
$r_{\mathrm{bonus}}$
& \makecell[c]{$\mathbbm{1}_{\mathrm{contact}}\mathbbm{1}_\mathrm{success}\frac{5.0}{1+10d_t}$ \\ $+\mathbbm{1}_\mathrm{stay\_success}\times10$} 
& 1.0
& 1.0
\\
$r_{\mathrm{curl}}$
& $-0.001\lVert\bm{\theta}_t\rVert_2^2$
& 1.0
& 10.0
\\
$r_{\mathrm{table}}$
& $\min(10\min_j(h^{\mathrm{finger}}_{j} - 0.62), 0)$
& 1.0
& 1.0
\\
$r_{\mathrm{action}}$
& $-0.01\lVert a_t\rVert_2^2$
& 1.0
& 5.0
\\
\bottomrule
\end{tabular}
}
\caption{Reward shaping for RL teacher training in different stages.}
\label{tab:reward}
\end{table}

\section{Domain Randomization and External Perturbation}

Isaac Gym~\cite{makoviychuk2021isaac} offers a suite of domain randomization functions for RL training. We detail our domain randomization setup in Table~\ref{tab:dr}.

In addition to physics parameter randomization, we also perform a random object force pushing mechanism to improve the policy's robustness to external perturbation. Specifically, we implement the pushing as linear and angular velocities applied to the object. Every four seconds in the simulation, we apply a linear xy velocity $\sim \mathcal{U}(-0.2, 0.2)$ and an angular velocity $\sim \mathcal{U}(-0.2, 0.2)$ to the object.

\begin{table}[!t]
\renewcommand\arraystretch{1.05}
\centering
\begin{tabular}{lccc}
\toprule
\textbf{Term}
& \textbf{Operation}
& \textbf{Distribution}
& \textbf{Range}
\\
\midrule
Observation White Noise 
& Additive
& Gaussian
& $[0, 0.002]$
\\
Observation Correlated Noise 
& Additive
& Gaussian
& $[0, 0.001]$
\\
\midrule
Action White Noise 
& Additive
& Gaussian
& $[0, 0.05]$
\\
Action Correlated Noise 
& Additive
& Gaussian
& $[0, 0.015]$
\\
\midrule
Gravity
& Additive
& Gaussian
& $[0, 0.4]$
\\
\midrule
Joint Stiffness
& Scaling
& Uniform
& $[0.9, 1.1]$
\\
Joint Damping
& Scaling
& Uniform
& $[0.9, 1.1]$
\\
\midrule
Joint Lower Limit
& Additive
& Gaussian
& $[0, 0.01]$
\\
Joint Upper Limit
& Additive
& Gaussian
& $[0, 0.01]$
\\
\midrule
Robot Mass
& Scaling
& Uniform
& $[0.5, 1.5]$
\\
Robot Friction
& Scaling
& Uniform
& $[0.7, 1.3]$
\\
\midrule
Object Mass
& Scaling
& Uniform
& $[0.5, 1.5]$
\\
Object Friction
& Scaling
& Uniform
& $[0.7, 1.3]$
\\
\bottomrule
\end{tabular}
\caption{Domain Randomization Setup.}
\label{tab:dr}
\end{table}

\section{Paired Point Masking}
\label{sec:app-masking}

As in Sec.~\ref{sec:teacher-student}, we introduce a \textbf{random plane-height masking strategy} for the student policy learning to improve our policy's robustness to real-world point input.

In student learning, for each environment, we first perform \textbf{plane masking}. Specifically, we randomly sample one plane that crosses the object's centroid, select one side of it, and then mask out all the points on that side. In this way, we obtain approximately half of the original object points. These point indices are kept the same throughout the whole environment. This process is to simulate the single-view observation in our real-world deployment so that our policy can generalize to varying camera views.

For the remaining points, at each timestep, we then apply \textbf{height masking}. Specifically, we first randomly sample a height ratio in $[0.2, 1.0]$, and based on the height, we mask out 90\% of the points above this height and 5\% of the points below. Finally, we apply a Gaussian noise $\sim \mathcal{N}(0,0.005)$ to the remaining points. This process is to simulate the occlusion between fingers and object points so that our policy can generalize to fewer object points and point noises in the real world.

\section{Dexterity on In-Hand and Non-Prehensile Manipulation}

\begin{figure}[t]
\centering
\includegraphics[width=1.0\linewidth]{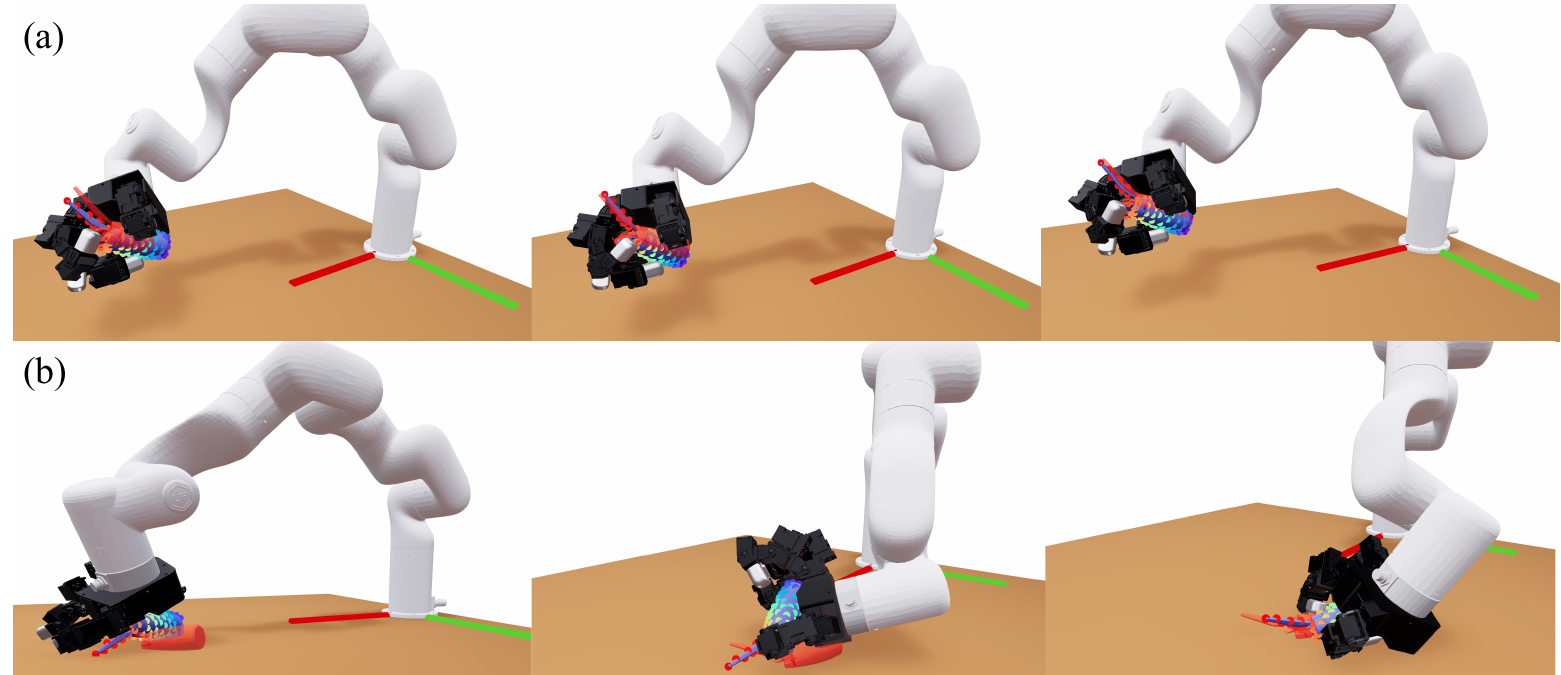}
  \caption{Our \method{} policy shows dexterity on in-hand and non-prehensile manipulation.}
  \label{fig:dexterity}
\end{figure}

Beyond prehensile manipulation, our \method{} also shows emergent dexterity on in-hand and non-prehensile manipulation learned in RL training. As shown in Fig.~\ref{fig:dexterity}(a) our AP2AP policy shows in-hand reorientation behaviors that rotate the object without moving the arm. In Fig.~\ref{fig:dexterity}(b), when the object accidentally falls off the hand onto the table, the policy could drive the finger to push the object and flick it around to reach the goal pose, showing evident non-prehensile capabilities. The results highlight that finger dexterity could emerge from our AP2AP training even when using low-cost hardware with limited dexterity.

\section{Cross-Embodiment Training}

\begin{figure}[t]
\centering
\includegraphics[width=1.0\linewidth]{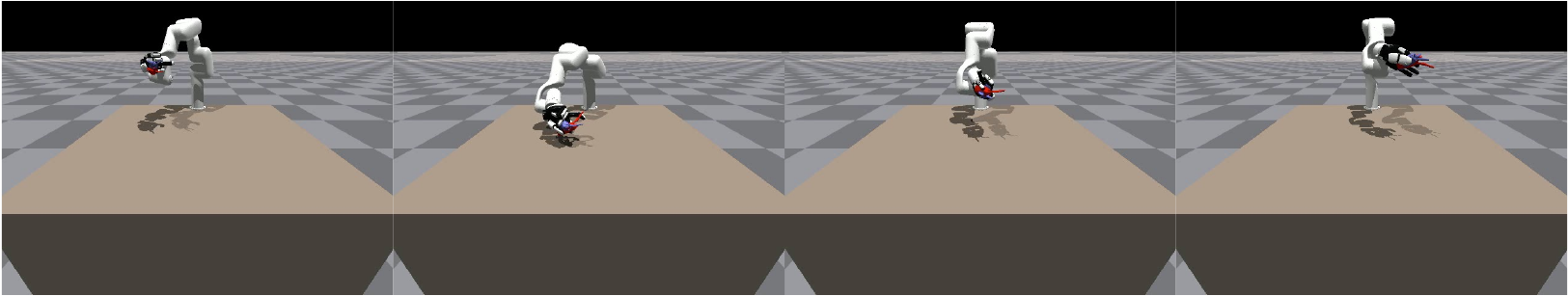}
  \caption{With the same simulation setting (reward, curriculum, etc.), the training can be seamlessly transferred to the XHand.}
  \label{fig:xhand}
\end{figure}

Our \method{} framework is a zero-shot sim-to-real method for dexterous manipulation, so one can update the robot URDF in the simulator for new embodiments seamlessly.
In Fig.~\ref{fig:xhand} we show that with the same simulation setting (rewards, curriculum, etc.), our AP2AP training can be seamlessly transferred to the XHand without any extra modifications.

\section{Training Stability}

\begin{figure}[t]
\centering
\includegraphics[width=0.5\linewidth]{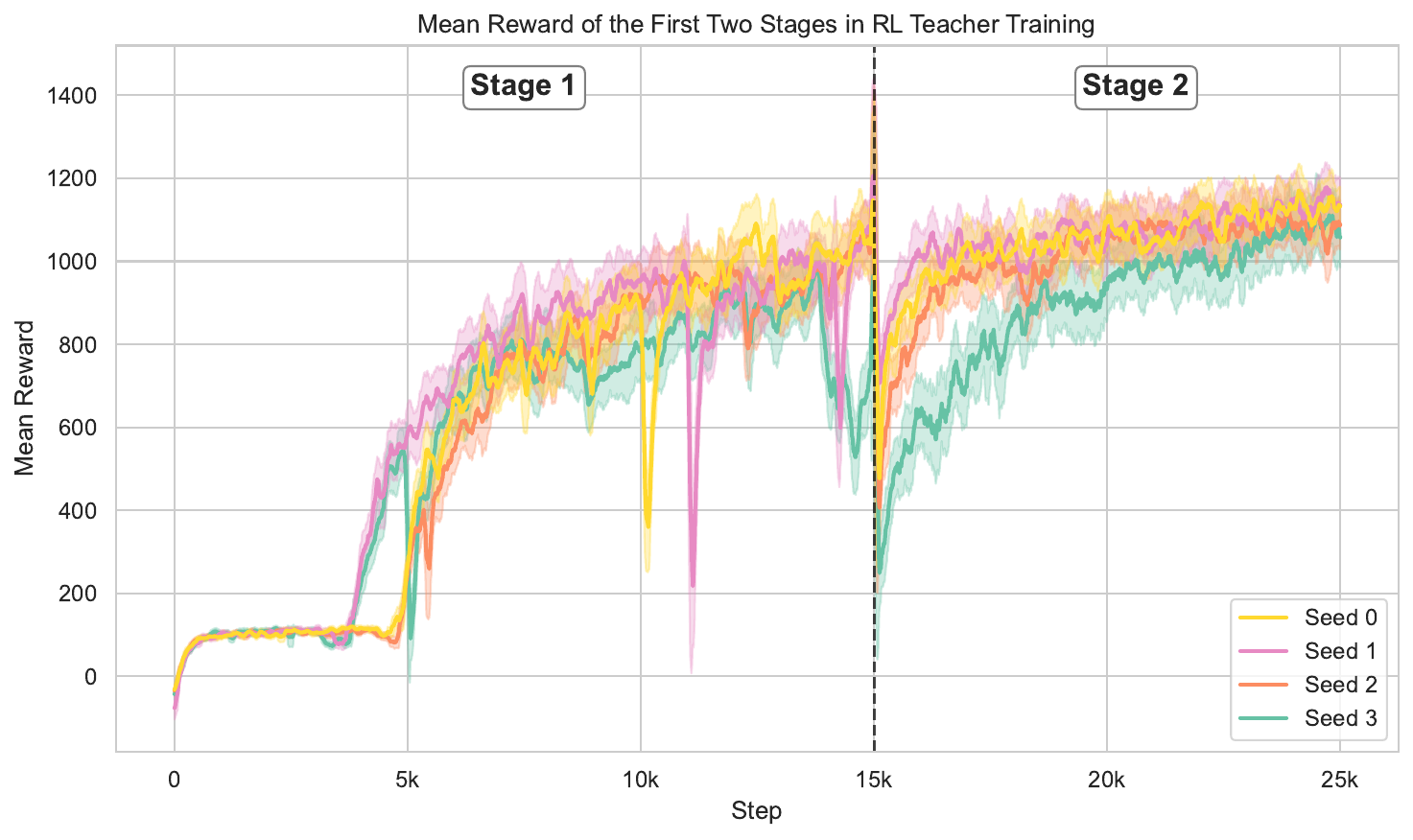}
  \caption{Our training is stable with different seeds.}
  \label{fig-reb:seed}
\end{figure}

We tried 3 additional random seeds for our teacher training, and we plot the curve in Fig.~\ref{fig-reb:seed}. The result shows that our training is stable with different seeds.

\section{Additional Hyperparameters}

\begin{table}[t]
\centering
\small
\setlength{\tabcolsep}{6pt}
\scalebox{1.0}{
\begin{tabular}{lc}
\toprule
\textbf{Hyperparameter}
& \textbf{Value}
\\
\midrule
Algorithm & PPO \\
Optimizer & Adam \\
Number of Environments & 4096 \\
Number of Object Points & 128 \\
Discount Factor ($\lambda$) & 0.95 \\
GAE parameter ($\gamma$) & 0.96 \\
Desired KL Divergence & 0.016 \\
Clip Range & 0.2 \\
Actor MLP Hidden Dimension & $[1024, 1024, 512, 512]$ \\
Critic MLP Hidden Dimension & $[1024, 1024, 512, 512]$ \\
PointNet Hidden Dimension & $[128, 128]$ \\
\bottomrule
\end{tabular}
}
\caption{Hyperparameters of the Teacher Policy.}
\label{tab:teacher_hyper}
\end{table}
\begin{table}[t]
\centering
\small
\setlength{\tabcolsep}{6pt}
\scalebox{1.0}{
\begin{tabular}{lc}
\toprule
\textbf{Hyperparameter}
& \textbf{Value}
\\
\midrule
Algorithm & DAgger \\
Optimizer & Adam \\
Number of Environments & 3200 \\
Number of Object Points & 64 \\
Learning Rate & $1\times 10^{-4}$ \\
Token Dimension & 128 \\
Number of Self-Attention Layers & 4 \\
\bottomrule
\end{tabular}
}
\caption{Hyperparameters of the Student Policy.}
\label{tab:student_hyper}
\end{table}

Additional hyperparameters of the teacher and student policies are detailed in Table~\ref{tab:teacher_hyper} and Table~\ref{tab:student_hyper}, respectively.

\end{document}